\documentclass[english]{IEEEtran}
\usepackage[T1]{fontenc}
\usepackage[latin9]{inputenc}
\usepackage{amsmath}
\usepackage{graphicx}
\usepackage{subfigure}
\usepackage{amssymb}
\usepackage{babel}
\usepackage{multirow}
\usepackage{tabularx}
\usepackage[flushleft]{threeparttable}
\usepackage{bm}
\usepackage{booktabs}
\graphicspath{{figures/}}

\begin{document}
\title{Map Container: A Map-based Framework for Cooperative Perception}

\author{Kun Jiang$^{1}$, Yining Shi$^{1}$, Benny Wijaya$^{1}$, Mengmeng Yang$^{1}$, \\ Tuopu Wen$^{1}$, Zhongyang Xiao$^{2}$, and Diange Yang$^{1}$
\thanks{$^{1}$Kun Jiang, Yining Shi, Benny Wijaya, Mengmeng Yang, Tuopu Wen, and Diange Yang are with 
the School of Vehicle and Mobility, Tsinghua University, Beijing, 100084, China. 
\tt \{jiangkun, yangmm\_qh, ydg\}@mail.tsinghua.edu.cn, \{syn21,huangjq19,wtp18\}@mails.tsinghua.edu.cn}%

\thanks{$^{2}$Zhongyang Xiao is with Alibaba Group, Beijing, 100015, China.
\tt xiaozhongyang@yeah.net}%

}
\maketitle
\begin{abstract}
The idea of cooperative perception is to benefit from shared perception data between multiple vehicles and overcome the limitations of on-board sensors on single vehicle. However, the fusion of multi-vehicle information is still challenging due to inaccurate localization, limited communication bandwidth and ambiguous fusion. Past practices simplify the problem by placing a precise GNSS localization system, manually specify the number of connected vehicles and determine the fusion strategy. This paper proposes a map-based cooperative perception framework, named map container, to improve the accuracy and robustness of cooperative perception, which ultimately overcomes this problem. The concept 'Map Container' denotes that the map serves as the platform to transform all information into the map coordinate space automatically and incorporate different sources of information in a distributed fusion architecture. In the proposed map container, the GNSS signal and the matching relationship between sensor feature and map feature are considered to optimize the estimation of environment states. Evaluation on simulation dataset and real-vehicle platform result validates the effectiveness of the proposed method.
\end{abstract}


\section{Introduction}
\IEEEPARstart{P}{erception} is a paramount aspect of safety in autonomous driving, which is responsible for the understanding and cognition of the states of ego vehicle and the driving environment. However, the existing perception technology based on on-board sensors of one single vehicle suffers several core issues. One obvious shortcoming is the blind spot, which is often caused by occlusion and the limited sensing range of various sensors \cite{Myakinkov2019}. Another important issue of single-vehicle perception is the vulnerability to sensor errors. 

Cooperative perception is believed to have the potential to solve these issues and more such as traffic congestion as proposed in\cite{Thandavarayan2020}. The general idea is to share information between intelligent vehicles to improve perception range and accuracy via vehicle-to-vehicle (V2V) and vehicle-to-everything (V2X) communication protocol \cite{Liu2013,Kim2015,Llatser2019}. More and more research has shown that this method could provide the ability to "see-around-the-corner" as well as "see-through" \cite{Shimizu2019} which is beneficial to anticipating the moving vehicles outside of the ego's vehicle area of vision. Furthermore, the perception range of ego vehicle can be extended beyond-line-of-sight \cite{Li2013} via V2X data sharing. 

However, the current multi-vehicle cooperative perception has not been deployed to large-scale applications, and there are several bottlenecks that have not yet been resolved. A typical problem is that the bandwidth of the wireless networking is limited, and the delay is unstably high, and real-time information cannot be transmitted over long distances. In addition, in the dense scene of various connected vehicles, the perception information needs to be properly organized, and a reasonable fusion architecture is required. Finally, high-precision positioning equipment is generally not equipped in low-cost autonomous driving solutions, and inaccurate positioning will affect the fusion characteristics in the perceptual information alignment step of cooperative perception.

\begin{figure*}[t]
\centering
\includegraphics[width=1\linewidth]{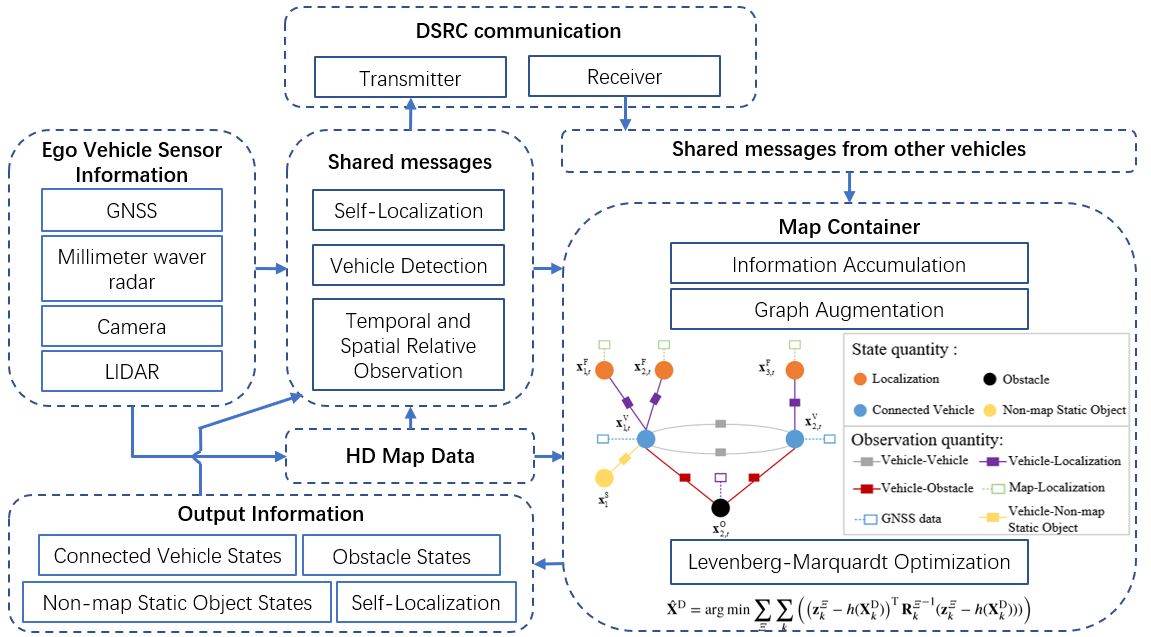}
\caption{Overall framework of map container. The map container accumulates information from ego vehicle and V2X, augments information in a factor graph, and solves the cross-localization fusion problem in a Levenberg-Marquardt optimization approach.}
\label{Flowchart}
\end{figure*}

In this paper, a new concept for cooperative perception, \textbf{map container}, is introduced. The overall framework of map container is shown in Fig. \ref{Flowchart}. In a typical cooperative perception scenario with several connect vehicles, traffic agents, and HD map. Connected vehicles are equipped with onboard sensors and transmit shared messages via Dedicated Short Range Communication (DSRC) communication. The perpectual outputs of interest are the localization states of ego and connected vehicles as well as other traffic agents.  The static information is sensorized and utilized by each connect vehicle in the form of a virtual \textbf{map sensor}. 

The container serves as an information gathering with reference to the map coordinate calculated. First, the image segmentation results are obtained from the monocular camera on the vehicle to perform the map matching process with the HD map, as previously published in \cite{Xiao2018}, to obtain the accurate vehicle localization. The map element information (lane lines, lamp post, and traffic signs) is also extracted from the map to serve as additional constraints in the joint state estimation process. In this container, all the information is unified into one coordinate system, making the state optimization process in the cooperative perception process practical. Furthermore, since perception information is used to obtained from accurate map information in the global coordinate system, the dependency of overlap of perception horizon between the connected vehicles can be eliminated. At the same time, the proposed approach can fully extend the perception ability of the ego vehicle to eliminate the blind spot area even though the connected vehicle is also in the blind spot area. 

In summary, the contributions of this paper can be summed up into three parts:
\begin{enumerate}
\item 
\textbf{A new concept framework of cooperative perception, Map Container}. HD map serves as both references coordinate and virtual sensor for base of cooperative perception. 

\item
\textbf{High precision self-localization and cross-localization with low-cost onboard sensors}. High localization precision is guaranteed by self-localization with monocular camera and gnss/imu equipments using a novel map-matching algorithm, and then refined by multiple observations by other connected vehicles of the ego vehicle. 

\item 
\textbf{A joint state estimation optimization process to estimate states in a unified manner}. Dynamic objects from ego and connected vehicles are estimated continuously in the form of trajectories. 

\end{enumerate}

The remainder of this paper is organized as follows: Section \ref{related_works} introduces the related works, including V2X-based cooperative perception, HD map localization and fusion architecture research. Section \ref{map_container} presents problem formulation and three key modules of map container: Distributed fusion architecture, map-matching localization, and driving space state estimator. Section \ref{sec:experiment} introduces the simulation and real-world experiments as well as their results and analyses. Section \ref{conclu} concludes the paper.

\section{Related Works}\label{related_works}
\subsection{V2X Cooperative Perception}

Cooperative perception is to utilize relative observations as correlations and redundancies between vehicle poses to estimate the ego vehicle's pose, as well as the targets' pose \cite{Li2012}. However, this idea often relies on the overlap of perception view between connected vehicle to accurately create correct correlations\cite{Kim2013}. This constraint usually hinders the application of such technology in the real driving scenario. However, in the real application, the error in acquiring perception information exists, making the correlation process difficult.

In addition,cooperative perception is still hindered by another issue that the sensor's information between vehicles does not have the same reference coordinate \cite{Soatti2018}. In recent years, researchers used the GNSS sensor result as the benchmark reference for all the information obtained by each of the connected vehicle \cite{Chong2013}. However, in complex scenes of real traffic scenario, GNSS information is often unstable, which can cause the drift in a certain direction for the perception result \cite{Lu2017}. This problem may deteriorate the state estimation result of the joint cooperative perception. 

In view of the lack of a unified test platform, experimental data and evaluation metrics in the early cooperative perception practice, a number of works in recent years have determined fair experimental data. Among them, in 2021, Runsheng Xu et al. \cite{OPV2V} and Yiming Li et al. \cite{V2X-Sim} used Carla and Sumo co-simulation to implement vehicle-road co-simulation platforms OPV2V and V2X-Sim respectively, and established a benchmark algorithm for cooperative object detection. Given rich training data, researchers start to focus on feature-level fusion between V2V or V2I. Feature fusion attempts include V2VNet proposed by Tsun-Hsuan Wang et al. \cite{V2VNet} that considered time compensation methods, and When2com \cite{When2com} and Who2com \cite{Who2com} networks proposed by Yen-Cheng Liu et al. respectively explored time compensation in collaboration and dynamic organization problem, Yiming Li et al. proposed DiscoNet \cite{DiscoNet} based on collaborative graph learning; Runsheng Xu et al. proposed a feature fusion network OPCOOD based on attention mechanism \cite{OPV2V}. However, experiments on simulation platforms ignored the influence of communication delay and positioning error. It is an unexplored issue how to realize the fusion of cooperative perception in a unified architecture when there are serious spatio-temporal errors in the real vehicle scene.

\subsection{HD Map Enhanced Perception} 

High definition (HD) map is a great source of information that can provide accurate static elements references to further improve the state estimation in cooperative perception \cite{Seif2016}. Unlike simultaneous localization and mapping (SLAM), where a map is generated simultaneously in real-time \cite{Alliez2020}, the HD map is assumed to be available in advance. The alignment between its feature landmarks can be done by detecting and matching from on-board sensors' perception to improve localization information. Over the past few years, the development of HD maps has been heavily linked with the development of V2X technology as it can provide high accuracy (centimetre-level precision) on self-vehicle localization. For example, the map used in \cite{Quack2017}  is built from Lidar sensor data and has a precision of up to 10 cm. Other researchers also have started to use this type of map as reference information \cite{Chong2013}. There is also research that integrates map-based localization with on-board vehicle sensors directly\cite{Atia2017}\cite{Oguz-Ekim2016}. In general, recent research use HD map information to improve the localization of self-vehicle\cite{Ma2019,Deng2019,Jeong2020,Pauls2020}. Since it is robust to the change in weather conditions which may influence the localization result, HD map is also used for other driving functions such as detection for 3D object\cite{yang2020}. Furthermore, the memory required for these map have been decreasing significantly with the implementation of vectorized data into the HD map information\cite{Kang2020}. Moreover, it avoids the alignment requirement of localization layer with other map features\cite{Hu2019}.

\subsection{Cooperative Perception Fusion Architecture}
Generally speaking, there are two types of cooperative perception fusion architecture, centralized cooperative perception and distributed cooperative perception. To the author's knowledge, at present, there is still no unified standard for cooperative perception architecture. But in general, they all need to satisfy the following requirements: 

\begin{enumerate}
\item Realizing the vehicle state estimation.
\item Obtaining the position of the road environment perception.
\item Making information interaction with driving environment.
\end{enumerate}

\textbf{Centralized Cooperative Perception} In the centered cooperative perception architecture, the cooperative perceptual container can obtain the perceptual data of all vehicles in the network, and solve the optimization problem centrally. In this architecture, the assume is that the autonomous vehicle can receive the original observations and map data of all other vehicles at the same time. And the optimization problem is directly solved in a centralized manner, then its Jacobian matrix is $\mathcal{M} * \mathcal{N}$ dimension, where $\mathcal{M}$ is the number of optimized states of the whole system and $\mathcal{N}$ is the number of observations of the whole system. When the number of connected cars in the scene increases, both $\mathcal{M}$ and $\mathcal{N}$  will increase correspondingly, which will lead to too much computation for the solution of self-vehicle centralized solution. In addition, the centered solution requires each vehicle to send all measurements directly, which will cause too much data transmission in the system \cite{Soatti2018}.

\textbf{Distributed Cooperative Perception} In the distributed architecture, each connected vehicle first estimates some state quantities based on some measurements of the local sensor system, and transmits the prior distribution information of these state variables, so as to avoid the transmission of the original measurements and reduce the amount of data transmitted by information in the IoV. In addition, the estimation results provide a better initial value for further state estimation at the center node and reduce the complexity of state estimation at the center node \cite{Shen2017}.

\section{Map Container}\label{map_container}

\subsection{Problem Formulation}
In this paper, cooperative perception is generally formulated as a state estimation problem, where the states are the foundation of the unified description of the environment, and the observation of states originate from ego-vehicle, V2X and HD map.

\textbf{Definition of States} The driving environment consists of two types of objects: dynamic obstacles, e.g. vehicles, pedestrians and cyclists, and static road features, e.g. lane lines and light poles. Then the states of a driving environment during a time window $T$ is defined as:
\begin{equation}
  \mathbf{X}^{\text{D}} = \{
    \mathbf{x}_{j,t_x}^{\text{O}},
    \mathbf{x}_{k,t_x}^{\text{R}},
     \}, t_x \in T
\end{equation}
where $\mathbf{x}_{i,t}^{\text{CO}}$ is the instant state of $i$th dynamic object, $\mathbf{x}_{j,t}^{\text{R}}$ is the instant state of $j$th road feature.

To be more specific,the state of a connected vehicle contains the vehicle position, attitude, speed and other attribute information, written as:
\begin{equation}
    \mathbf{x}_{i,t_x}^{\text{C}}
    =
      [
      x_{i,t_x},
      y_{i,t_x},
      z_{i,t_x},
      \theta_{i,t_x},
      \psi_{i,t_x},
      \phi_{i,t_x}
      ] ^ T
  \end{equation}
where $x_{i,t_x}$, $y_{i ,t_x}$ and $z_{i,t_x}$ are center of dynamic objects in the global coordinates; $\theta_t$, $\psi_t$ and $\phi_t$ are the corresponding pitch angle, yaw angle and roll angle.

the state of a road feature mainly contains: 
\begin{equation}
 \mathbf{x}_{o,t_x}^{\text{F}} = [
 x_{i,t_x},
 y_{i,t_x},
 z_{i,t_x}
]
\label{point_feature_equation}
 \end{equation}
 
Where $x_{i,t_x}$, $y_{i,t_x}$ and $z_{i,t_x}$ are the coordinate values of the target's reference point in the global coordinate.

\textbf{Definition of Observation} In an cooperative perception scenario, the perception sources include ego vehicle, connected vehicles with on-board units(OBU), road-side unit(RSU) and the HD map. The whole set of all observation data during time window $T$ is recorded as $\mathbf{Z}^{\text{P}}$.
\begin{equation}
  \mathbf{Z}^{\text{P}}= \left\{\mathbf{Z}^{\text{ego}},\mathbf{Z}^{\text{v2x}},\mathbf{Z}^{\text{map}}
  \right\} 
\end{equation}

\textbf{Definition of HD map} The HD map is an important source of observation information about the road environment, for example, the accurate position of lane-markings and road signs. Therefore, map is utilized as a virtual sensor named \textbf{map sensor} in this paper. Observation of map sensors consists of road features positions, the map-vehicle relative positions, and map-obstacle relative positions:
\begin{equation}
\begin{array}{c}\mathrm{Z}^{\mathrm{M}}=\left\{\mathbf{Z}^{\mathrm{M}-\mathrm{F}}, \mathrm{Z}^{\mathrm{M}-\mathrm{V}}, \mathrm{Z}^{\mathrm{M}-\mathrm{O}}\right\} =\left\{\mathbf{z}_{j,f}^{\mathrm{M}-\mathrm{F}}, \mathbf{z}_{j, v}^{\mathrm{M}-\mathrm{V}}, \mathbf{z}_{j, o}^{\mathrm{M}-\mathrm{O}}\right\}\end{array}
\label{map_measuremtn}
\end{equation}
The prior information about the location feature in the map is $\mathbf{z}_{j,f}^{\text{M-F}}$, where $f$ is the index of road feature; Similarly, the relative information about connected vehicles and the map feature is  defined as $\mathbf{z}_{j,v}^{\text{M-V}}$;The relative information about the  obstacles and the map feature is defined as $\mathbf{z}_{j,o}^{\text{M-O}}$.

Map features are classified as point, line, and surface features. For example, a light pole is linear feature controlled by two reference points, and a lane line is a surface feature. To process different feature types, a map matching model is developed in Section \ref{Map-matching}.

\textbf{Definition of cooperative perception problem solved by map Container} Assuming that the intelligent and connected vehicles can obtain the asynchronous time set of observations in space $\mathbf{Z}^{\text{P}}$. The driving environment space $\mathbf{X}^{\text{D}}$ is based on unified expression in the global coordinate, in which state quantities estimate the self-localization and perception of other features at the current moment $\hat{\mathbf{X}}_{t_n}$. The optimization problem is to use redundant observation information as constraints and obtain an accurate state quantity by optimizing the residual error between observation and state.

\subsection{Distributed Fusion Architecture}
As reviewed in Section \ref{related_works}, centralized architecture can surely obtain a high-precision perception result. However, in practical application, this architecture puts too much pressure on information transmission and computing resources in the Internet of Vehicles (IoV). In addition, a better initial value of the state variables should be provided when the state estimation problem is solved in an optimal way. While in the centralized state estimation, it is hard for the original measurement information transmitted by each vehicle without processing to provide a good initial value.

\begin{figure*}[ht]
  \centering
  \includegraphics[scale=0.5]{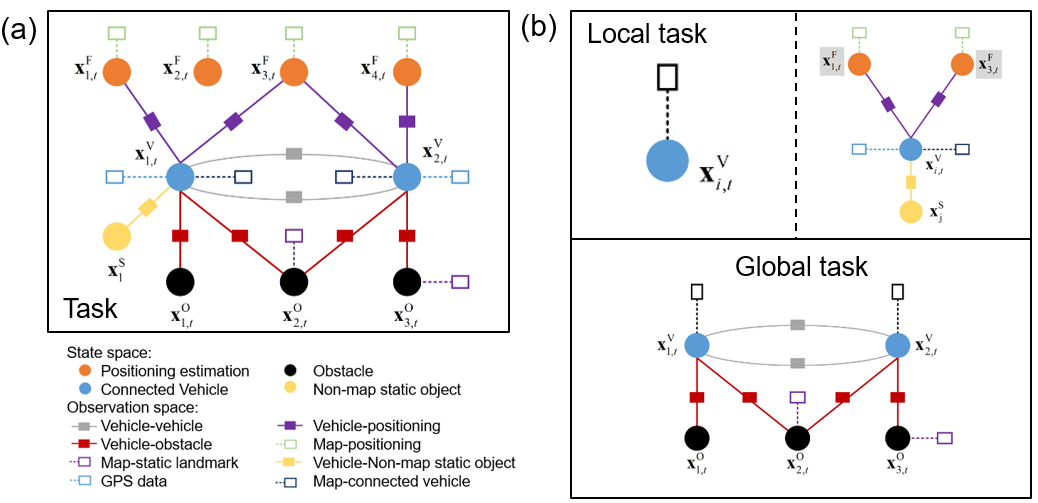}
  \caption{The task division of different cooperative perception architecture: (a) Centralized cooperative perception architecture. (b) Distributed cooperative perception architecture.}
  \label{Coop_arch}
\end{figure*}

As shown in Fig. \ref{Coop_arch}, a distributed cooperative perception framework is adopted in this paper to solve the cooperative perception. The optimization problem is split into three parts. At first each intelligent car estimates part of the state variables according to local observations, and then partial observation information and estimation results are transmitted to the host vehicle. Finally, the host car solves part of the global state, so as to reduce the computational complexity of a single node. At the same time, in the distributed solution mode, each vehicle sends the pre-estimated results and part of the observation information, so as to avoid the direct transmission of all the observations and reduce the overall data transmission.

\subsection{Map-matching Localization}\label{Map-matching}
Spatial alignment of multi-source observation from on-board sensors, HD map or V2X is the fundamental step of cooperative perception. In contrast to former practices using GPS and IMU localization approaches, a fundamental benefit of the proposed map container technology is the ability to transform multi-source information into map coordinate with high precision automatically via a map-matching approach. 

\begin{figure}[ht]
  \centering
  \includegraphics[scale=0.5]{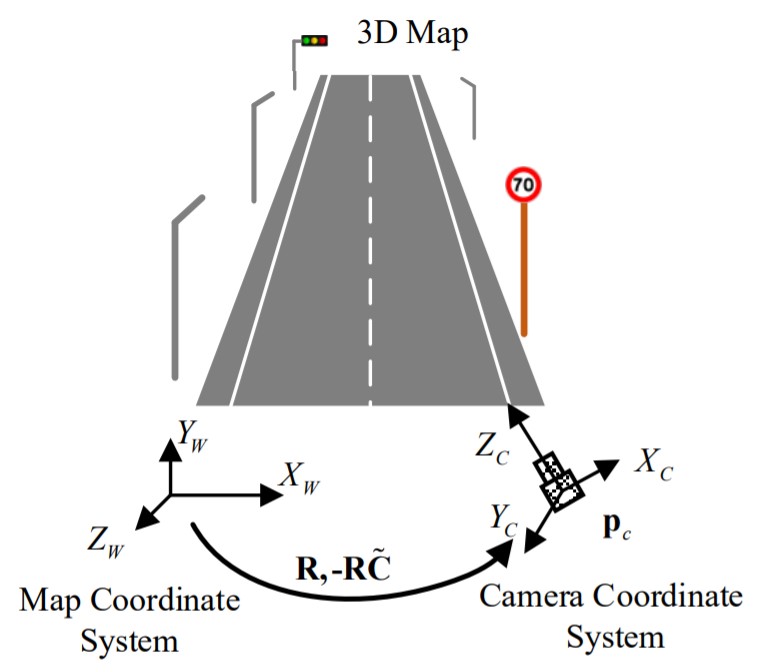}
  \caption{An graphical illustration of the camera pose estimation with the help of HD map.}
  \label{Camera_pose_estimation}
\end{figure}

The matching process between monocular camera and HD map is introduced, which comes down to a 3D-2D (map to image view) registration problem. An overview of the proposed map-matching method is shown in figure \ref{Camera_pose_estimation}. The pre-selected landmarks from the 3D map are projected according to a supposed camera pose, and a cost function is calculated considering the re-projection error of these landmarks and their aligned features on the captured image. The camera pose is estimated iteratively by minimizing the cost function with the Levenberg-Marquardt algorithm.

A 6-DOF camera pose model is modeled as $\boldsymbol{p}_{\mathrm{c}}=\left(C_{x}, C_{y}, C_{z}, \phi, \theta, \psi\right)^{T}$ to describe the camera pose, where $C_{x}$,$C_{y}$,$C_{z}$ are the three coordinates of the camera center in the map coordinate system, and $\phi$, $\theta$, $\psi$ are the yaw, pitch, and roll angle of the camera coordinate system relative to the map coordinate system. A finite projective camera model is used to define the camera projection function $\Pi:\left(\mathbb{R}^{3}, \mathbb{R}^{6}\right) \rightarrow \mathbb{R}^{2}$ that projects the 3D homogeneous point $\mathrm{X}=(\mathrm{X}, \mathrm{Y}, \mathrm{Z}, 1)^{\mathrm{T}}$ in the map coordinate to the 2D image coordinate $\mathbf{u}=(u, v, 1)^{T}$ known the 6-DOF camera pose $\mathbf{p}_{\mathrm{c}}$:
\begin{equation}
    \Pi: \mathbf{u}=\mathbf{K} \mathbf{R}[\mathbf{I} \mid-\tilde{\mathbf{C}}] \mathbf{X}
\end{equation}

where $K$ is the internal parameters of the camera, $\mid-\tilde{\mathbf{C}}]$ is the coordinate of the camera center in the map coordinate system, and $R$ is the relative rotation matrix from the map coordinate system to the camera coordinate system. Based on $\Pi$ , A point projection function is defined as $\pi_{\mathrm{P}}:\left(\mathbb{R}^{3}, \mathbb{R}^{6}\right) \rightarrow \mathbb{R}^{2}$ and a line projection function $\pi_{L}:\left(\mathbb{R}^{6}, \mathbb{R}^{6}\right) \rightarrow \mathbb{R}^{4}$ projects the 3D point and line features in the map to the image domain given the camera pose $\mathbf{p}_{\mathrm{c}}$:
\begin{equation}
    \begin{array}{l}\mathrm{u}=\pi_{\mathrm{P}}\left(\mathbf{p}_{m}, \mathbf{p}_{c}\right) \\\mathrm{l}=\pi_{\mathrm{L}}\left(\mathbf{l}_{m}, \mathbf{p}_{c}\right)\end{array}
\end{equation}

\begin{figure}[ht]
  \centering
  \includegraphics[scale=0.83]{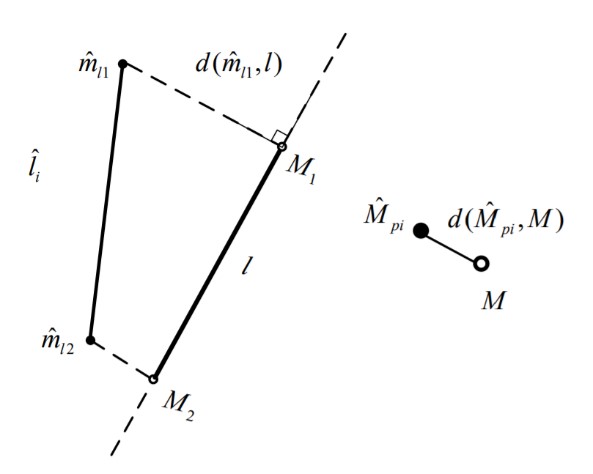}
  \caption{An graphical illustration of the definition of the distance between lines and points.}
  \label{definition_of_distance_camera}
\end{figure}

where $\mathbf{p}_{\mathrm{m}} \in \mathbb{R}^{3}$ denotes the center of a point landmark, and $\mathbf{l}_{\mathrm{m}} \in \mathbb{R}^{6}$ denotes the line landmark. The two control points in $\mathbf{I}_{\mathrm{m}}$ is respectively projected to the image coordinate according to the projecting matrix. $\mathbf{l}=\left(u_{1}^{T}, u_{2}^{T}\right)^{T} \in \mathbb{R}^{4}$ are used to describe the projected line landmark in the image coordinate system. 
The basic idea of estimating the camera pose is to maximize the similarity between the projected features and the current image frame. A residual model is proposed based on both lines and points features' distance, given the extracted map $\mathbf{m}_{\mathrm{e}}$, the recognized lines and points features $\mathbf{L}_{\mathrm{r}}$,$\mathbf{P}_{\mathrm{r}}$ with their association $c$:

\begin{equation}
\begin{array}{r}\mathrm{r}\left(\mathbf{m}_{\mathrm{e}}, \mathbf{L}_{\mathbf{r}}, \mathbf{P}_{\mathbf{r}}, \mathbf{c}, \mathbf{p}_{\mathbf{c}}\right)=\sum_{c_{i} \in \boldsymbol{c}_{l}} \boldsymbol{D}_{L}^{2}\left(\boldsymbol{\pi}_{L}\left(\boldsymbol{L}_{\boldsymbol{m}}^{i}, \boldsymbol{p}_{\boldsymbol{c}}\right), \boldsymbol{l}_{\boldsymbol{c}_{l}^{i}}\right) \\+\sum_{\boldsymbol{c}_{P}^{i} \in \boldsymbol{c}_{p}} \boldsymbol{D}_{P}^{2}\left(\boldsymbol{\pi}_{P}\left(\boldsymbol{P}_{\boldsymbol{m}}^{i}, \boldsymbol{p}_{c}\right), \boldsymbol{p}_{c_{l}^{i}}\right)\end{array}
\end{equation}

Here the correspondence set $\mathrm{c}=\left(\mathbf{c}_{\mathbf{l}}, \mathbf{c}_{\mathbf{p}}\right)$ of features is defined on the map and the current image, where the line correspondence $\mathrm{c}_{\mathrm{l}}=\left\{c_{l}^{1}, c_{l}^{2}, \ldots\right\}, c_{i} \in\{1, \ldots, N\}$, where $N$ is the number of line features on the current image), and $\mathrm{c}_{l}^{i}=j \leq N$ if the $i$-th line feature in the pre-selected landmarks correspond to the $j$-th line feature on the current image. The point correspondence set $c_{\mathrm{p}}$ is defined in the same way. 

The definition of the distance between two lines $D_{L}:\left(\mathbb{R}^{4}, \mathbb{R}^{4}\right) \rightarrow \mathbb{R}$ in the image coordinate system is explained in figure \ref{definition_of_distance_camera} is a projected line from a 3D feature in the map, and $\hat{l}_{i} \in \mathbb{R}^{4}$ is an extracted line in the current image with two endpoints $M_{1}$ and $M_{2}$. The distance of $\hat{l}_{i}$ and $l$ is defined by the following equation: 

\begin{equation}
D_{L}\left(\hat{l}_{i}, l\right)=\frac{1}{2} \sum_{i=1}^{2} \frac{\left\|M_{1}-M_{2} \mid \widehat{m}_{l i}-M_{1}\right\|}{\left\|M_{1}-M_{2}\right\|_{2}}
\end{equation}

where, $\mathrm{m}_{\mathrm{l} 1}$ and $\mathrm{m}_{\mathrm{l} 2}$ are the projected point of the two 3D line feature endpoints. And the distance of two aligned points is calculated with $\mathrm{D}_{\mathrm{p}}:\left(\mathbb{R}^{2}, \mathbb{R}^{2}\right) \rightarrow \mathbb{R}$:
\begin{equation}
    D_{P}\left(\widehat{M}_{p i}, M\right)=\left\|\widehat{M}_{p i}-M\right\|_{2}
\end{equation}

where $\widehat{M}_{p j}$ is the projected point landmark and $M$ is its corresponding point recognized in the current frame. In the urban driving scene, the vehicle seldom leaves the ground, and the ground is flat in most cases. To improve the convergence of the optimization, especially inparse landmark case, a soft constraint is added to pitch, roll angle of the camera $\theta$, $\psi$ and the height of the camera $C_{y}$, by adding one item to the cost function: 
\begin{equation}
    \mathrm{r}_{\mathrm{n}}\left(\mathbf{p}_{c}\right)=\left\|\left(\theta, \psi, C_{y}-\left(y_{\text {lane }}+H\right)\right)\right\|_{2}^{2}
\end{equation}

where, $y_{\text {lane }}$ is the height of the nearest control point of lanes extracted from the map, $H$ is the height of the camera off the ground. Denote the whole residual as: 
\begin{equation}
    \begin{array}{c}\mathbf{R}\left(\boldsymbol{p}_{\mathrm{c}}\right)=\sum_{c_{i}^{j} \in c_{l}} D_{L}^{2}\left(\pi_{L}\left(L_{m}^{i}, p_{c}\right), l_{c_{l}^{i}}\right) \\+\sum_{c_{i}^{j} \in c_{l}} D_{P}^{2}\left(\pi_{P}\left(\mathbf{P}_{m}^{i}, \mathbf{p}_{c}\right), \mathbf{p}_{c_{l}^{i}}\right) \\+\lambda_{n}^{2}\left\|\left(\theta, \psi, C_{y}-\left(y_{l a n e}+H\right)\right)\right\|_{2}^{2}\end{array}
\end{equation}

where $\lambda_{n}$ is the weight of the added item in the overall cost function, which is set to 0.001 in the test settings (the length is in centimeters and the angle is in angular degree). The optimal estimation of the camera pose relative to the map can be obtained by solving the optimization problem. The problem is addressed with Levenberg-Marquart method:
\begin{equation}
    \mathrm{p}_{\mathrm{c}}^{*}=\underset{P_{c}}{\operatorname{argmin}}\left[\mathbf{r}\left(\mathbf{m}_{\mathbf{e}}, \mathbf{L}_{\mathbf{r}}, \mathbf{P}_{\mathbf{r}}, \mathbf{c}, \mathbf{p}_{\mathbf{c}}\right)+\lambda_{\mathrm{n}}^{2} r_{n}\left(\mathbf{p}_{c}\right)\right]
\end{equation}





The control points in the domain-like features are expressed in the predicted value:
\begin{equation}
    \mathbf{h}_{r}^{\mathrm{R}}\left(\mathbf{m}_{r}^{\mathrm{R}}, \mathbf{x}_{k}\right)=\left[\begin{array}{c}\mathbf{h}_{r, 1}^{\mathrm{R}} \\\cdots \\\mathbf{h}_{r, N^{\mathrm{R}}}^{\mathrm{R}}\end{array}\right]=\left[\begin{array}{c}\pi\left(\mathbf{p}_{r, 1}^{\mathrm{L}}, \mathbf{x}_{k}\right) \\\cdots \\\pi\left(\mathbf{p}_{r, N^{\mathrm{R}}}^{\mathrm{L}}, \mathbf{x}_{k}\right)\end{array}\right]
\end{equation}

\begin{figure}[ht]
  \centering
  \includegraphics[width=0.45\textwidth]{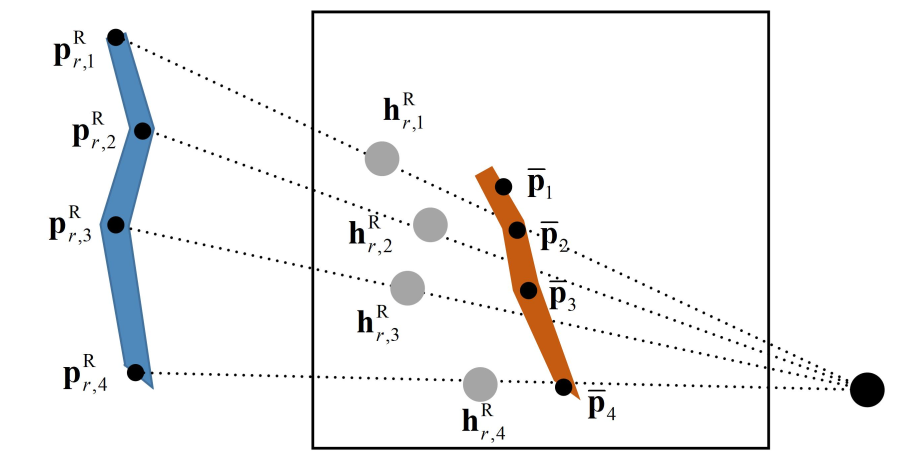}
  \caption{Schematic diagram of the definition of domain-like feature observation residuals}
  \label{plane_feature} 
\end{figure}

As shown in figure \ref{plane_feature}, assuming that each point in the predicted value of the domain feature is $\mathbf{h}_{r, i}^{\mathrm{R}}$,and the corresponding point identified in the image is $\overline{\mathbf{p}}_{i}$, then theoretically the residual of the domain feature is each re-projection The distance from the point to its corresponding recognition point is $\mathbf{h}_{r, i}^{\mathrm{R}} \overline{\mathbf{p}}_{i}$. However, it is impossible to accurately determine the exact position of the point $\overline{\mathbf{p}}_{i}$ from the domain features in the image, and it is also difficult to determine the association relationship between it and each re-projection point.

The Euclidean Distance Transform (EDT) method is used to convert the point-to-point distance in the residual to the value of the re-projection point in the transformation map. Define a grid $\mathcal{G}$ whose size is the same as the image $\mathbf{I}_{k}$ used for feature recognition. Then the curve-like features extracted from the image are $\mathcal{R}_{k} \subset \mathcal{G}$ a point $\mathbf{p} \in \mathcal{G}$ in the grid, the distance of the feature $\mathcal{R}_{k}$ is transformed into $\mathcal{D}_{\mathcal{R}_{k}}(\mathbf{p}): \mathbb{R}^{2} \in \mathcal{G} \rightarrow \mathbb{R}$. Define the distance transformation map $I_{\mathcal{R}_{k}}$ about the domain-like feature $\mathcal{R}_{k}$, and each pixel takes the value of the distance transformation value about the feature $\mathcal{R}_{k}$:





The value of the re-projection point $\mathbf{h}_{r, i}^{\mathrm{R}}$ on the distance transformation graph $\mathcal{I}_{\mathcal{R}_{k}}$ approximates the residual $\mathbf{h}_{r, i}^{\mathrm{R}} \overline{\mathbf{p}}_{i}$ in figure \ref{plane_feature}. The constrained residual of a domain-like feature $\mathbf{m}_{r}^{\mathrm{R}}$ is defined as:

\begin{equation}
    \mathbf{r}^{\mathcal{R}}\left(\mathcal{R}_{k}, \mathbf{m}_{r}^{\mathrm{R}}, \mathbf{x}_{k}\right)=\left[\begin{array}{c}\mathcal{I}_{\mathcal{R}_{k}}\left(\pi\left(\mathbf{p}_{r, 1}^{\mathrm{R}}, \mathbf{x}_{k}\right)\right) \\\vdots \\\mathcal{I}_{\mathcal{R}_{k}}\left(\pi\left(\mathbf{p}_{r, N^{\mathrm{R}}}^{\mathrm{R}}, \mathbf{x}_{k}\right)\right)\end{array}\right]
\end{equation}

It should be noted that, for the lane line feature, in order to distinguish between the linear model and the domain model, in the visual image processing, the number of exterior points after fitting is calculated. When the ratio of the number of external points to the number of data points exceeds a certain threshold $T^{\text {Outlier }}$, the residual error is described by the domain model, otherwise the linear model is used. When calculating the residuals of the domain features, it is not necessary to specify the association relationship between the recognition feature and the location feature in the map like point and line features. In principle, the association method of domain-like features is the nearest matching in the image coordinate. In the process of constructing the distance transformation map, the process of map-camera feature matching has been implicitly completed.

\subsection{Driving Environment Space State Estimator}

Now that the observation has been precisely transformed into the map coordinate, the state estimation is formulated as an optimization problem.

According to the Maximum Likelihood Estimation (MLE) theory, the state estimator maximizes the probability of the observations conditioned on the state set:
\begin{equation}
  \hat{\mathbf{X}}^{\text{D}} =\mathop{\arg \max}_{\mathbf{X}^{\text{D}}} P\left(\mathbf{Z }^{\text{P}}|\mathbf{X}^{\text{D}}\right)
\label{equ:MLE_prolbem}
\end{equation}

$ \hat{\mathbf{X}}^{\text{D}} $ is the estimated result of the driving environment space state parameters in large likelihood, $P\left(\mathbf{Z}^{ \text{P}}|\mathbf{X}^{\text{D}}\right)$ is the given state parameter $\mathbf{X}^{\text{D}}$, the set of observations $  \mathbf{Z}^{\text{P}}$.

Considering the process of generating observation data from a certain sensor, the observation $\mathbf{z}_k$ is a consequence of the state $\mathbf{X}^{\text{D}}_{k}$ and a random error.  For a certain observation value $\mathbf{z}_k$ in the cooperative sensing system, the observation equation is expressed using measurement equation:
\begin{equation}
  \label{equ:system_mearement}
  \mathbf{z}_k = h(\mathbf{X}^{\text{D}}_k) + \mathbf{v}_k
\end{equation}
$h(\cdot)$ is a function of state quantity, which describes the predicted value of the observation data generated by the sensor in the system state $\mathbf{X}^{\text{D}}_k$. $ \mathbf{v}_k$ is the measurement noise, which reflects the uncertainty in the process of sensor production of observation data. Under the assumption of Gaussian noise,
\begin{equation}
 \mathbf{v}_k \sim \mathcal{N}(\mathbf{0}, \mathbf{R}_k)
\end{equation}
 $\mathbf{R}_k$ is the co-variance matrix of noise $ \mathbf{v}_k$.

Further, consider the observations at different moments. Assuming that the measurement probabilities of each sensing unit and each time in the observation are independent of each other, the likelihood probability density function of the system can be obtained from the factor graph as all sensors in the system's perception space for a while. The likelihood of the conditional probabilities of all observations in $\mathcal{K}$ which is the product of all conditional probabilities:

\begin{equation}
  \begin{split}
    \label{equ:MLE_problem_expanded}
    P\left(
      \mathbf{Z}^{\text{P}}|\mathbf{X}^{\text{D}} 
      \right)
      &= P \left(\mathbf{Z}^{\text{V}}, \mathbf{Z}^{\text{M}},\mathbf{Z}^{\text{G}} | \mathbf{X}^{\text{D}} \right)
     =  \prod_{\Xi}    \prod_ k P\left(\mathbf{z}_k^{\Xi}|\mathbf{X}^{\text{D}}\right) , \\
     &
    \Xi \in \{\text{V-O,S,V,F}\} \cup \{ \text{M-O,V,F} \} \cup \{ \text{G}\}, k \in \mathcal{K}
  \end{split}
\end{equation}

Then the probability distribution of the observed value of any single sensor in Eq. \ref{equ:MLE_problem_expanded} at time $k$ is:
\begin{equation}
\begin{array}{c}P\left(\mathbf{z}_{k}^{\Xi} \mid X^{\mathrm{D}}\right)=\mathcal{N}\left(h(\mathbf{X}^{\text{D}}_k)), \mathbf{R}_{k}^{\Xi}\right) \\=\frac{1}{\sqrt{(2 \pi)^{N} \operatorname{det}\left(\mathbf{R}_{k}^{\Xi}\right)}} \exp \left(-\frac{1}{2}\left(\mathbf{z}_{k}^{\Xi}-h(\mathbf{X}^{\text{D}}_k)\right)^{\mathrm{T}}\right. \\\left.{\mathbf{R}^{\Xi}_k }^{-1}  \left(\mathbf{z}_{k}^{\Xi}-h(\mathbf{X}^{\text{D}}_k)\right)\right)\end{array}
\end{equation}

Bring the above formula into Eq. \ref{equ:MLE_problem_expanded}), and then take the negative logarithm on both sides to get
\begin{equation}
  \begin{split}
    &- \ln(P(\mathbf{Z}^{P}|\mathbf{X}^{D})) = -\sum_{\Xi} \sum_{k} \ln \left( (2 \pi)^N \det (\mathbf{R}_k^\Xi)
    \right)
    \\
    &+ \frac{1}{2}\sum_{\Xi} \sum_{k}  \left(    \left(\mathbf{z}_k^{\Xi} - h(\mathbf{X}^{\text{D}}_k)\right) ^\mathrm{T} \right.
    \left.{\mathbf{R}^{\Xi}_k }^{-1}  \left(\mathbf{z}_k^{\Xi} - h(\mathbf{X}^{\text{D}}_k)\right) \right)
  \end{split}
\end{equation}

Considering that the first term on the right is a constant, the optimization problem of solving formula (\ref{equ:MLE_prolbem}) can be summarized as:

\begin{equation}
\begin{array}{r}\hat{\mathbf{X}}^{\mathrm{D}}=\arg \min \sum_{\Xi}\left.\sum_{k}\left(\left(\mathbf{z}_{k}^{\Xi}-h(\mathbf{X}^{\text{D}}_k)\right)^{\mathrm{T}}\right.\right. \\\left.\left.{\mathbf{R}^{\Xi}_k}^{-1}  \left(\mathbf{z}_{k}^{\Xi}-h(\mathbf{X}^{\text{D}}_k)\right)\right)\right)\end{array}
\end{equation}
    
Introduce the definition of residual of measurements here:
\begin{equation}
  \label{equ:uniform_residule}
  \mathbf{e}_k^{\Xi}=\mathbf{z}_k^{\Xi}-h(\mathbf{X}^{\text{D}}_k)
\end{equation}

It reflects the difference between the actual observation value $\mathbf{z}_k^{\Xi} $ and the predicted value $h_a$ of a certain sensor, and Eq. \ref{equ:MLE_prolbem} is equivalent to adjusting the state quantity ${\mathbf{X}^{D}}$ makes the residual term of each observation in the sense of ${\mathbf{R}_k^{\Xi} }^{-1}$ at every moment in the perceptual space The sum of the following norms is the smallest:
\begin{equation}
  \label{equ_residula_general}
  {\hat{\mathbf{X}}^{\text{D}}}
  = \arg \min \sum_{\Xi}
  \sum_{k}{ {\mathbf{e} }_k^{\Xi} }^\mathrm{T} {\mathbf{R}^{\Xi}_k }^{-1}
    \mathbf{e}_k^{\Xi}
\end{equation}

Eq. \ref{equ_residula_general} uses the inverse matrix of the covariance matrix of each sensor noise ${\mathbf{R}^{\Xi}_k }^{-1} $ to adjust the sensor data of different accuracy. 

Levenberg-Marquardt method is adopted to solve the optimization problem. The residuals of each constraint, defined in formula (\ref{equ:uniform_residule}), are arranged together in a column vector to form a nonlinear target:
\begin{equation}
f\left(\mathbf{X}_{k}^{\mathrm{D}}\right)=\left[\begin{array}{c}\mathbf{r}^{\mathrm{V}-\mathrm{V}} \\\mathbf{r}^{\mathrm{V}-\mathrm{r}} \\\vdots \\\mathbf{r}^{\mathrm{V}-0}\end{array}\right]=\left[\begin{array}{c}\mathbf{r}_{1} \\\mathbf{r}_{2} \\\vdots \\\mathbf{r}_{\mathrm{n}}\end{array}\right]
\end{equation}
The optimization target is to get the minimum residuals, that is, to make the inner product of the column vectors composed of the residuals as small as possible, so the problem turns into a least-squares problem.
\begin{equation}
\hat{\mathbf{X}_{k}}^{\mathrm{D}}=\arg \min f\left(\mathbf{X}_{k}^{\mathrm{D}}\right)^{\mathrm{T}} f\left(\mathbf{X}_{k}^{\mathrm{D}}\right)
\end{equation}

Given an initial value $\hat{\mathbf{X}}_{0}^{\mathrm{D}}$, the Levenberg-Marquardt iteration is executed until the estimated state converges.
\begin{equation}
\left(\mathrm{J}\left(\mathrm{x}_{k}\right)^{\mathrm{T}} \mathrm{J}\left(\mathrm{x}_{k}\right)+\lambda \mathrm{I}\right) \Delta \mathrm{x}_{k}=-\mathrm{J}\left(\mathrm{x}_{k}\right)^{\mathrm{T}} f\left(\mathrm{x}_{k}\right)
\end{equation}

\begin{equation}
\mathbf{X}_{\mathrm{k+1}}^{\mathrm{D}}=\mathbf{X}_{\mathrm{k}}^{\mathrm{D}}+\Delta \mathbf{X}_{\mathrm{k}}
\label{jacobi_matrix}
\end{equation}

In formula (\ref{jacobi_matrix}), the Jacobi matrix size determines the computational complexity.

\section{Experiments}\label{sec:experiment}
\subsection{Simulation Validation}
\subsubsection{Simulation Setup} 
To verify the effectiveness of the cooperative perception architecture in a typical scenario, a simulation experiment is set up on the VISSIM platform. The configuration for the scenario and each measurements are as follows: 4 connected vehicles, 22 static map features (14 lamps, 4 traffic lights, and 4 traffic signs), 14 moving obstacles (8 pedestrians and 6 vehicles). In this scenario, the experiment was repeated 200 times, and the noise was added to the observation independently each time. 
\subsubsection{Evaluation Metrics} 
To quantitatively evaluate the localization performance, the Root Mean Square Error (RMSE) of the estimated states against the reference states is adopted as main evaluation metric. The RMSE is calculated as:
    \begin{equation}
    e_{\text{RMSE},i,t}^{\text{V}} =
    \sqrt{\frac{1}{M N^\text{V}}
    \sum_{j=1}^M \sum_{i=1}^{N^\text{V}} \begin{Vmatrix}
    \hat{\mathbf{p}}^{\text{V}}_{i,j,t}-
    \mathbf{p}^{\text{V}}_{i,j,t}
    \end{Vmatrix}^2 _2}
    \label{rms1}
    \end{equation}
    
Furthermore, to evaluate the positioning performance of other targets, the target localization result is also compared with the ground truth data in the vehicle coordinate system. The target positioning accuracy is calculated by RMSE of all targets within the perception area range.
    \begin{equation}
    e_{\text{RMSE},k,t}^{\text{T}} =
    \sqrt{\frac{1}{N^\text{O}M}
    \sum_{{k=1}^N}^\text{O} \sum_{j=1}^{M} \begin{Vmatrix}
    \hat{\mathbf{p}}^{\text{T}}_{k,j,t}-
    \mathbf{p}^{\text{T}}_{k,j,t}
    \end{Vmatrix}^2 _2}
    \label{rms2}
    \end{equation}

where, $M$ is the number of experiments, $p_(i,j,t)^V$ is the positioning result of car i in simulation $j$ at time $t$, and $p_(i,j,t)^V$ is the corresponding positioning ground truth values. $p_(k,j,t)^T$ is the positioning result of the $k$ target in the vehicle coordinate system in simulation $j$ at time $t$, and $p_(k,j,t)^T$ is its corresponding ground-truth value. 
\subsubsection{Experiment and Analysis}
\begin{figure}[ht]
\centering
\includegraphics[width=0.75\linewidth]{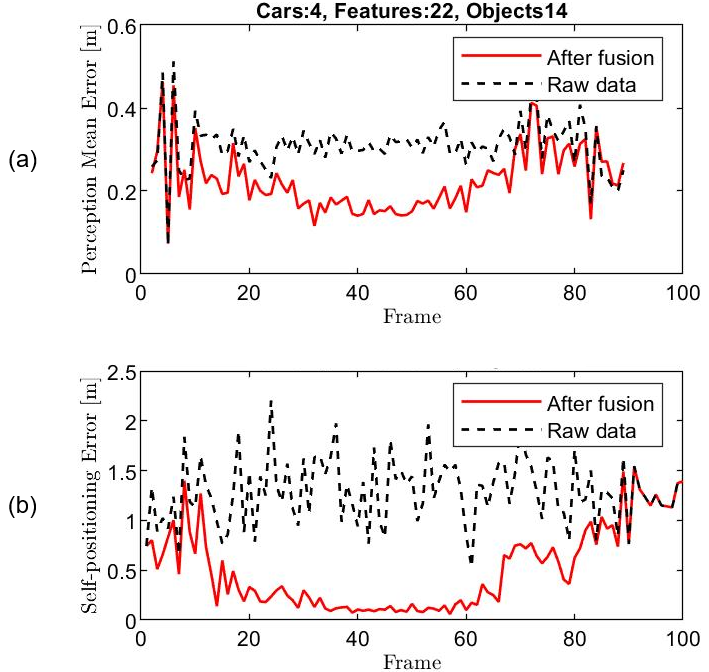}
\caption{Evaluation of simulation analysis: (a) RMSE of ego vehicle self-positioning in urban and suburban scenarios. (b) RMSE of targets relative-positioning in urban and suburban scenarios.}
\label{fig:exp_sim}
\end{figure}

Fig. \ref{fig:exp_sim} shows quantitative evaluation of simulation analysis. The result shows that the positioning accuracy of the proposed architecture for V2X technology is superior compared to the GNSS data for self-localization. The accuracy in urban areas is better when compared with the suburban areas because the dense map features in urban areas give a great improvement for map matching in particular.
    
When the target localization accuracy is considered, the proposed V2X architecture is superior to the ego-vehicle on-board sensor. This is because the information sent from other connected vehicles helps to reduce the uncertainty in the state estimation of the target. Furthermore, similar to the previous analysis, the urban areas seem to reduce the error in target localization. This is also because when the area is filled with dense map features, which can help improve the state estimation for the target.

\subsection{Test Vehicle Evaluation}

\subsubsection{Scenario and Vehicle Setup}

\begin{figure}[ht]
\centering
\includegraphics[width=0.7\linewidth]{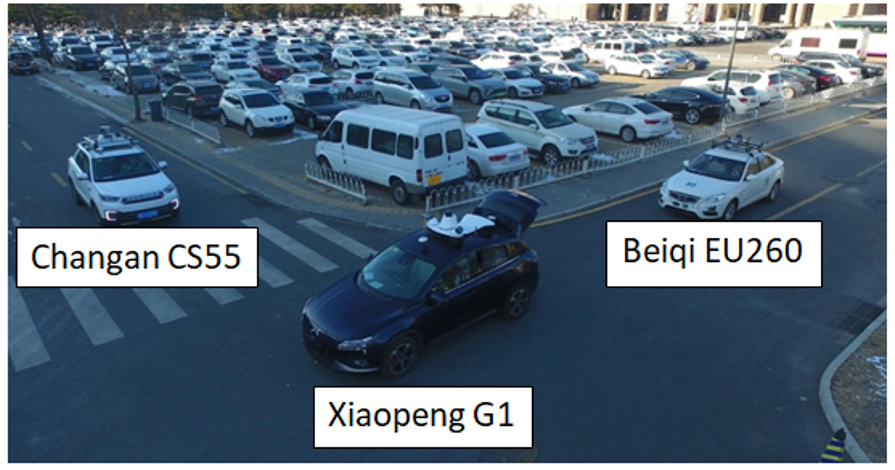}
\caption{Multi-vehicle cooperative perception test platform.}
\label{fig:test_platform}
\end{figure}

\textbf{Role settings} A multi-vehicle experiment platform is designed with three intelligent vehicles in real traffic scenarios: BAIC EU260, XiaoPeng G1, and Changan CS55. These vehicles can be seen in Fig. \ref{fig:test_platform}. One vehicle is acting as ego vehicle, the second vehicle is the connected vehicle, and finally, the third vehicle as to the moving obstacle. the moving obstacle is put within the perception range of the ego and connected vehicle. These experiments are conducted to show the perception improvement and extension by using the proposed cooperative perception framework. 

\begin{figure}[ht]
\centering
\includegraphics[width=0.75\linewidth]{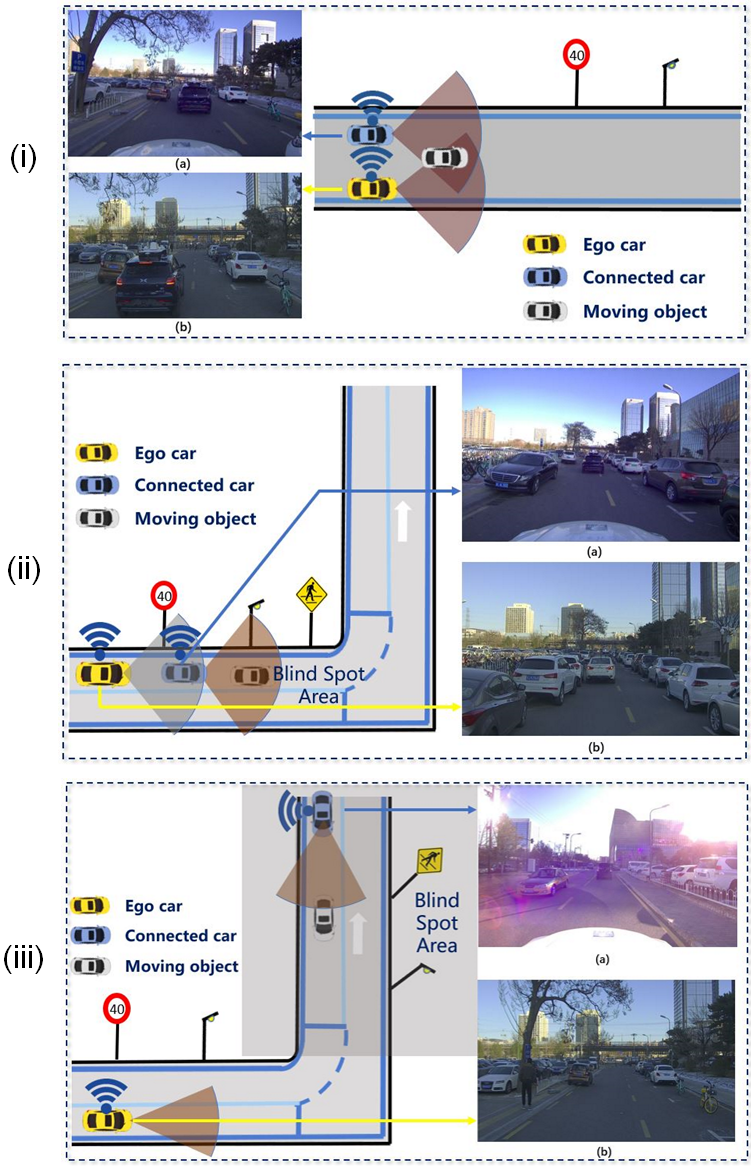}
\caption{A graphic diagram of the setup of cooperative perception in the real vehicle experiment where (a) is the vision captured from the connected vehicle (b) is the vision captured from ego vehicle. (i) Is the perception overlap experiment where the perception result is enhanced. (ii) and (iii) Are the perception extension experiment where the ego vehicle are not able to detect the target directly.}
\label{fig:exp}
\end{figure} 

\textbf{Scenarios setup} The first scenario, shown in (i) at Fig.  \ref{fig:exp}, is a typical scenario when the target vehicle is within the perception range of both ego and connected vehicles. The second scenario, shown in (ii) at Fig. \ref{fig:exp}, shows the extension of the perception range of the ego vehicle to detect a moving obstacle in the blind spot area, where the first part is when only the connected vehicle is within the perception range of the ego vehicle. The third scenario, shown in (iii) at Fig.  \ref{fig:exp} is when both of the connected vehicle and the moving target are not in the field of vision of the ego vehicle perception range.

\textbf{Vehicles and sensors setup} A  Three vehicles are utilized to conduct three types of experiments in different scenarios. Sensor data are from a monocular camera, vectorized HD map, millimetre-wave radar, GNSS, IMU data and Dedicated Short Range Communication (DSRC) equipments, the details of which are shown in Table. \ref{tab:sensors}. 

	






    \begin{table*}[t!]
    	\caption{
    		Sensors equipped on the vehicles
    	}
    	
    	\label{tab:sensors}
    	\centering
    	\begin{tabular}{ l c c c }
    		\toprule
    		\multirow{2}{*}{\textbf{Sensor Type}} & \multicolumn{3}{c}{\textbf{Sensor Name}}   \\ 
    		\cmidrule{2-4}  &\textbf{BAIC EU260}  &\textbf{XiaoPeng G1} &\textbf{Changan CS55}   \\ 
    		\midrule
    
    		\textbf{Camera} & Basler acA1920-40gc  &  Leopard LI-AR0231-AP0200-GMSL2 & Leopard USB30   \\
    
    		\textbf{Lidar} & Velodyne HDL-32E & RS-Lidar-32 & Velodyne HDL-64E  \\
    
    		\textbf{Millimeter Radar} & Continental ASR408  & Continental ASR408  & Continental ASR408  \\
    		\textbf{RTK-GNSS/IMU} & Trimble BD982/OXTS Inertial+ & PP7D-EI & Trimble BD982/OXTS Inertial+            \\
    		\textbf{GPS sensor}  & Ublox NEO-6M  &Ublox NEO-6M  &Ublox NEO-6M \\
            \textbf{Controller} &MXC-6401D/M4G & Nuvo-6108GC &MXC-6401D/M4G \\
    		\textbf{DSRC}  & MOKAR V2X OBE & MOKAR V2X OBE & MOKAR V2X OBE \\
    		\bottomrule
    	\end{tabular} 
    
    \end{table*}

\subsubsection{Perception Overlap Between Two Connected Vehicles}

\begin{figure}[ht]
\centering
\includegraphics[width=0.65\linewidth]{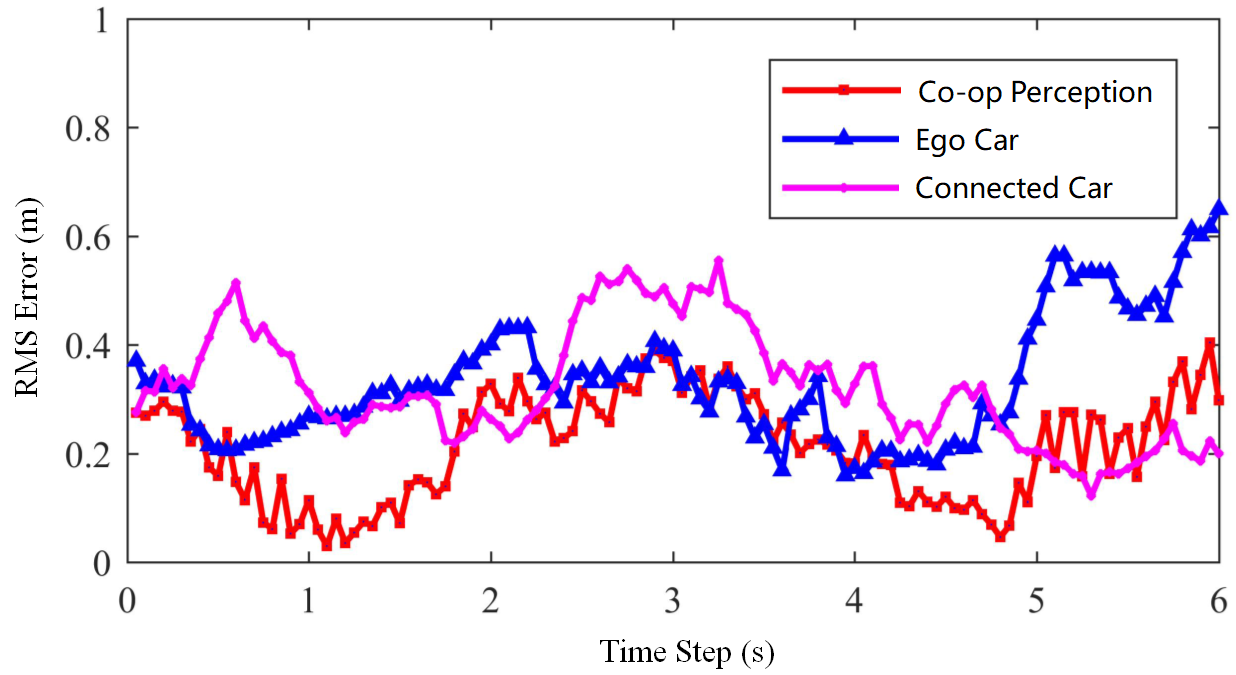}
\caption{RMS error distribution and it shows that cooperative perception yields a better accuracy compared to ego car only perception result}
\label{fig:res1}
\end{figure}

The quantitative evaluation of scenario (i) is illustrated in Fig.  \ref{fig:res1}. In scenario (i) at Fig. \ref{fig:exp}, The advantage of using a map container is analyzed when both connected vehicle and ego vehicle have a direct line of sight on the moving obstacle. Each experiment is performed two times to ensure the consistency of the result. Since the information is shared between connected vehicle, it means that there are redundant information about the moving obstacle received on the ego vehicle. This information can enhance the state estimation of the moving obstacle by using the proposed method.

Table. \ref{tab:res1} shows that the proposed method has a significant improvement compared to single vehicle perception result, by reducing the state error by about 14\% to 34\%. This signifies that when redundant information serves as reasonable constraints, the predicted state outcome becomes more accurate.
 
\begin{table}[ht]
 \centering
 \caption{Results of state estimation of the moving obstacle in cases where perception between connected vehicle are in the overlapped states}
 \label{tab:res1}
 \begin{threeparttable}
 \def\arraystretch{1.2}
 \begin{tabular}{cccc}
     \hline
      & Ego Car (m)  & Connected Car (m)  & Proposed Method (m) \\ 
     \hline
     1 & 0.33 & 0.35 & 0.23 \\
     2  & 0.37 & 0.34 & 0.29 \\

     \hline

 \end{tabular}
 \begin{tablenotes}
     \item *The result shown are calculated based on average RMS error of the moving obstacle states given the information from each vehicle input. 
 \end{tablenotes}
 \end{threeparttable}
 \end{table}

\subsubsection{Extension of Perception Area of Ego Vehicle}
\begin{figure}[ht]
\centering
\includegraphics[width=0.7\linewidth]{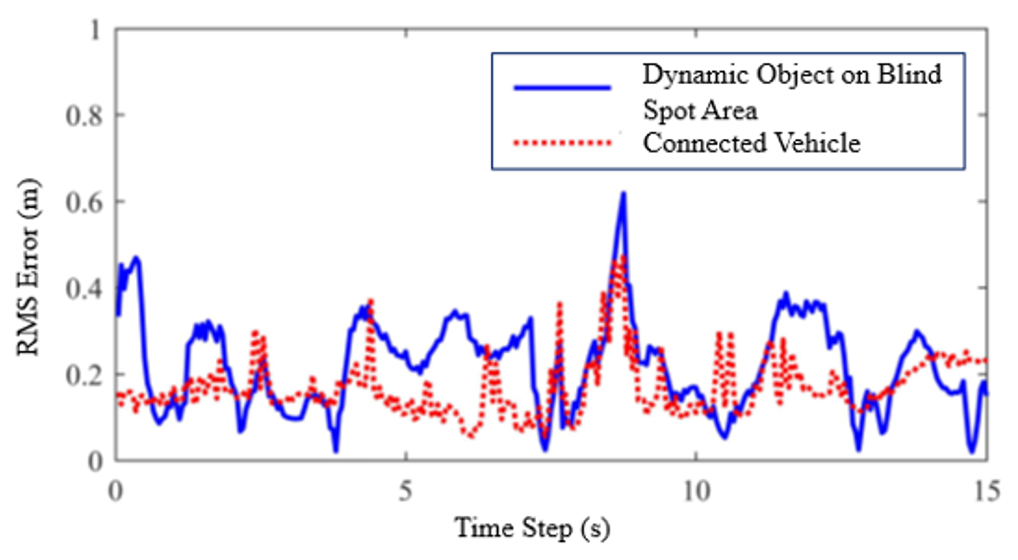}
\caption{RMS error for connected vehicle and dynamic object positioning accuracy for the first experiment of extension of perception area}
\label{fig:res2}
\end{figure}

\begin{figure}[ht]
\centering
\includegraphics[width=0.7\linewidth]{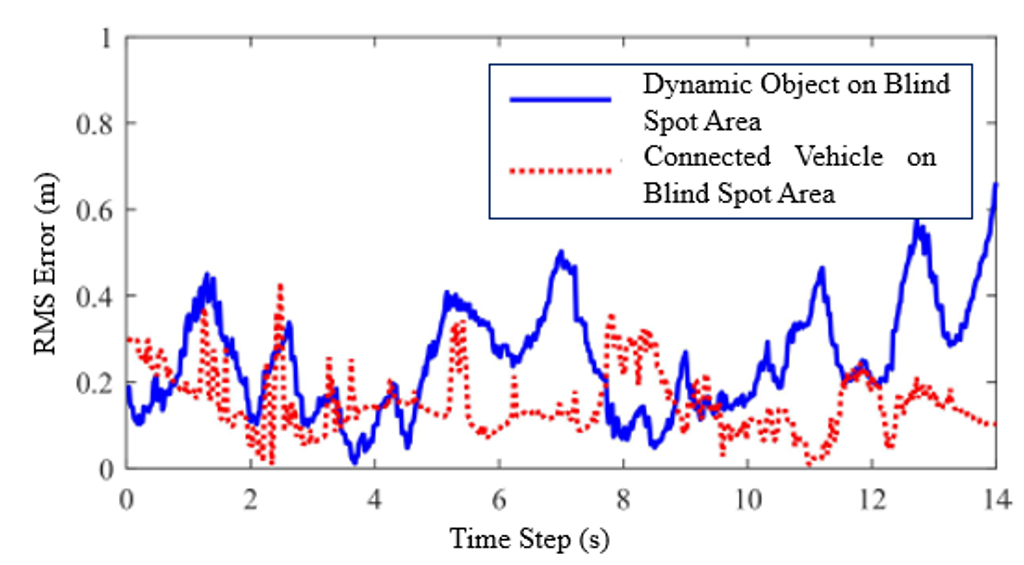}
\caption{RMSE error for connected vehicle and dynamic object positioning accuracy for the second experiment of extension of perception area}
\label{fig:res3}

\end{figure}

The quantitative evaluation of scenario (ii) and (iii) is illustrated respectively in Fig. \ref{fig:res2} and Fig. \ref{fig:res3}. In scenario (ii) and (iii) at Fig. \ref{fig:exp}, the perception area is extended to the blind spot area. In the first scenario, the ego vehicle is driving from west to east in the southern section of the test site, and there is a connected vehicle in front of it, so the ego vehicle's view is blocked by the connected vehicle and forms a blind zone as shown in Fig. \ref{fig:exp}. 

In the first experiment, the root means square of the positioning error for the connected car is 0.18m, and the root means square of the positioning error for the dynamic obstacle is 0.29m. It can be seen that in scenes where blind spots are generated due to occlusion, cooperative perception by using a map container can eliminate blind spot area by only using on-board sensor data combined with V2X.

In the second scenario, the ego vehicle is driving from west to east in the southern section of the test site, and the dynamic obstacle is driving in front of the connected vehicle as shown in Fig. \ref{fig:exp}. It can be seen that the connected vehicle and dynamic obstacles are in the blind spot area, which is outside the perception range of the ego vehicle. During this test, the ego vehicle can only rely on cooperative perception to detect the dynamic obstacle, thus expanding the perception range of the ego vehicle. 

Table. \ref{tab:res2} shows that the root mean square error of perception for connected vehicles is 0.18m, and the root mean square error of perception for the dynamic obstacle is 0.33m. Finally, it can be seen that the sensing accuracy of dynamic targets in the blind zone based on the cooperative perception system reaches the decimeter level in both of the experiments.

\begin{table}[ht]
\centering
\caption{Results of state estimation of the connected car and moving obstacle in blind spot cases.}
\label{tab:res2}
\begin{threeparttable}
\def\arraystretch{1.2}

\begin{tabular}{ccc}
 \hline
 Experiment &  Connected Car (m)  &  Dynamic Obstacle (m) \\ 
 \hline
 1 & 0.18 & 0.29 \\
 2 & 0.18 & 0.33 \\
 \hline

\end{tabular}
\begin{tablenotes}
 \item *The result shown are calculated based on average RMS error of the moving obstacle states given the information from each vehicle input. 
\end{tablenotes}
\end{threeparttable}
\end{table}

\section{CONCLUSION}\label{conclu}
This paper presents map container, a novel approach for cooperative perception using a HD map virtual sensor and a map container. HD map is used to provide high-precision benchmarks for spatio-temporal transformations of perceptual information. The proposed method demonstrates a robust fusion architecture to apply the information received directly into the joint state estimation optimization process that can organize perceptual messages vital to ego vehicle's driving space. The proposed method is evaluated in a simulation designed in the VISSIM simulator and real vehicle experiments. Both of the results show that the map container improves the state estimation of the targets. Even in cases where a blind spot exists, the proposed method is able to extend the perception range of the ego vehicle while still maintaining a decimeter level of accuracy.

\section*{ACKNOWLEDGEMENT}
This work was supported by National Natural Science Foundation of China (61903220), in part by the National Natural Science Foundation of China (U1864203).

\bibliographystyle{IEEEtran}
\bibliography{ref/ref}

\end{document}